\documentclass[10pt,twocolumn,letterpaper]{article}

\usepackage{wacv}
\usepackage{times}
\usepackage{epsfig}
\usepackage{graphicx}
\usepackage{amsmath}
\usepackage{amssymb}
\usepackage{cleveref}
\newcommand{\PaperName}{MLRDSC}
\newcommand{\sm}[2]{\scaleto{#1\mathstrut}{#2pt}}
\newcommand{\tav}[1]{\scaleto{#1\mathstrut}{6pt}}
\newcommand{\tavv}[1]{\scaleto{#1\mathstrut}{5pt}}
\newcommand{\tavvv}[1]{\scaleto{#1\mathstrut}{6.5pt}}
\newcommand{\msh}{\!\!\:}

\newcommand{\Dbf}{\mathbf{D}}
\newcommand{\Frm}{\mathrm{F}}
\newcommand{\Ibf}{\mathbf{I}}
\newcommand{\Qbf}{\mathbf{Q}}
\newcommand{\Xbf}{\mathbf{X}}
\newcommand{\Wbf}{\mathbf{W}}
\newcommand{\Zbf}{\mathbf{Z}}
\newcommand{\xbf}{\mathbf{x}}

\newcommand{\Cbf}{\mathbf{C}}
\newcommand{\Rbb}{\mathbb{R}}
\newcommand{\Sbb}{\mathbb{S}}

\newcommand{\Ecal}{\mathcal{E}}
\newcommand{\Dcal}{\mathcal{D}}

\newcommand{\Lcal}{\mathcal{L}}

\newcommand{\astaccent}[1]{\accentset{\ast}{#1}}
\crefname{figure}{Figure}{Figures}
\crefname{equation}{Equation}{Equations}
\crefname{algorithm}{Algorithm}{Algorithms}
\crefmultiformat{equation}{(#2#1#3)}{\,--\,(#2#1#3)}{#2#1#3}{#2#1#3}
\crefname{table}{Table}{Tables}
\usepackage{amsfonts,mathrsfs,bm,amsthm} 
\usepackage{pdfpages}
\usepackage{subcaption}
\usepackage{makecell}
\usepackage{enumitem}
\usepackage{tabularx}
\usepackage{algorithm}
\usepackage{algpseudocode}
\usepackage{url}
\usepackage{multirow}
\usepackage{rotating}
\usepackage{array}
\usepackage{caption}
\usepackage{pgfplots}
\usepackage{accents}
\usepackage{scalerel}
\usepackage{tikz}
\usepackage{booktabs}

\wacvfinalcopy 

\def\wacvPaperID{986} 

\ifwacvfinal\fi
\begin{document}

\title{Multi-Level Representation Learning for Deep Subspace Clustering}

\author{Mohsen Kheirandishfard \hspace{1.2cm} Fariba Zohrizadeh \hspace{1.2cm} Farhad Kamangar\\
Department of Computer Science and Engineering,\\
University of Texas at Arlington, USA\\
{\tt\small \{mohsen.kheirandishfard,fariba.zohrizadeh\}@gmail.com\hspace{.7cm}kamangar@cse.uta.edu}
}

\maketitle
\ifwacvfinal\thispagestyle{empty}\fi

\begin{abstract}
This paper proposes a novel deep subspace clustering approach which uses convolutional autoencoders to transform input images into new representations lying on a union of linear subspaces. The first contribution of our work is to insert multiple fully-connected linear layers between the encoder layers and their corresponding decoder layers to promote learning more favorable representations for subspace clustering. These connection layers facilitate the feature learning procedure by combining low-level and high-level information for generating multiple sets of self-expressive and informative representations at different levels of the encoder. Moreover, we introduce a novel loss minimization problem which leverages an initial clustering of the samples to effectively fuse the multi-level representations and recover the underlying subspaces more accurately. The loss function is then minimized through an iterative scheme which alternatively updates the network parameters and produces new clusterings of the samples. Experiments on four real-world datasets demonstrate that our approach exhibits superior performance compared to the state-of-the-art methods on most of the subspace clustering problems.
\end{abstract}

\section{Introduction}
Subspace clustering is an unsupervised learning task with a variety of machine learning applications such as motion segmentation \cite{lauer2009spectral,rao2010motion}, face clustering \cite{cao2015diversity,zhang2015low}, movie recommendation \cite{ntoutsi2014strength,zhang2012guess}, etc. The primary goal of this task is to partition a set of data samples, drawn from a union of low-dimensional subspaces, into disjoint clusters such that the samples within each cluster belong to the same subspace \cite{agrawal1998automatic,parsons2004subspace}. A large body of subspace clustering literature relies on the concept of self-expressiveness which states that each sample point in a union of subspaces is efficiently expressible in terms of a linear (or affine) combination of other points in the subspaces \cite{elhamifar2013sparse}. Given that, it is expected that the nonzero coefficients in the linear representation of each sample correspond to the points of the same subspace as the given sample. In order to successfully infer such underlying relationships among the samples and to partition them into their respective subspaces, a common practice approach is to first learn an affinity matrix from the input data and then apply the spectral clustering technique \cite{ng2002spectral} to recover the clusters. Recently, these spectral clustering-based approaches have shown special interest in utilizing sparse or low-rank representations of the samples to create more accurate affinity matrices \cite{elhamifar2013sparse,favaro2011closed,liu2012robust,luo2018consistent,vidal2014low}. A well-established instance is sparse subspace clustering (SSC) \cite{elhamifar2013sparse} which uses an $\ell_1$-regularized model to select only a small subset of points belonging to the same subspace for reconstructing each data point. More theoretical and practical aspects of the SSC algorithm are investigated and studied in detail in \cite{peng2013scalable,soltanolkotabi2012geometric,you2016scalable,you2015geometric}.

Despite the key role that the self-expressiveness plays in the literature, it may not be satisfied in a wide range of applications in which samples lie on non-linear subspaces, e.g. face images taken under non-uniform illumination and at different poses \cite{ji2017deep}. A common practice technique to handle these cases is to leverage well-known kernel trick to implicitly map the samples into a higher dimensional space so that they better conform to linear subspaces \cite{patel2013latent,patel2014kernel,xiao2015robust,yin2016kernel}. Although this strategy has demonstrated empirical success, it is not widely applicable to various applications, mainly because identifying an appropriate kernel function for a given set of data points is a quite difficult task \cite{zhang2019neural}.

\begin{figure}[t!]
\subcaptionbox{}{
\begin{subfigure}[normal]{0.48\linewidth}
\centering
\includegraphics[width=\linewidth]{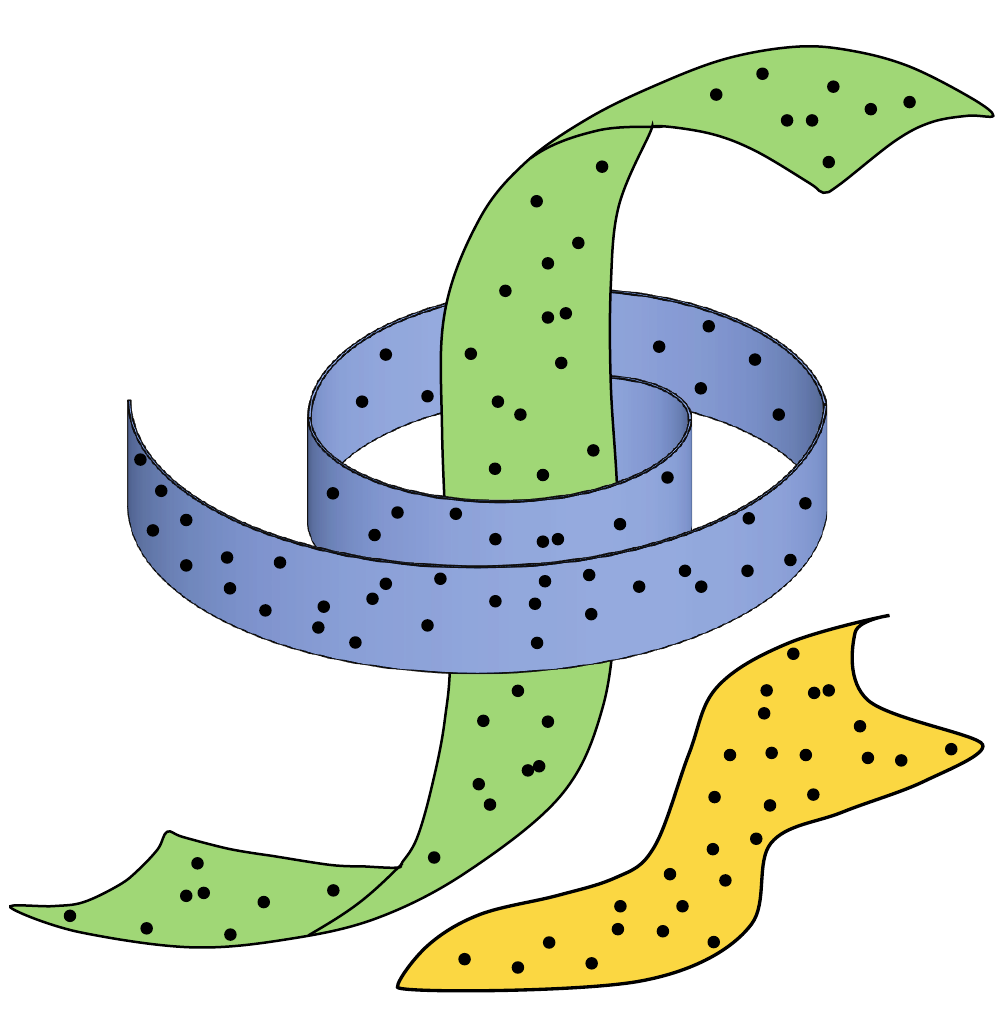}
\end{subfigure}}
\subcaptionbox{}{
\begin{subfigure}[normal]{0.48\linewidth}
\centering
\includegraphics[width=\linewidth]{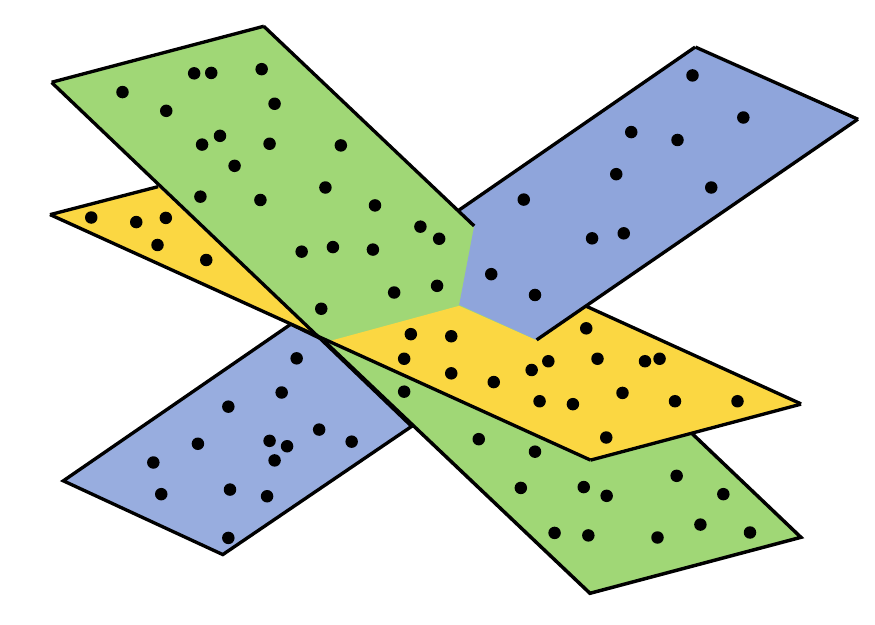}
\end{subfigure}}
\caption{Illustration of representation learning for subspace clustering. (a) Sample points may come from a union of nonlinear subspaces; (b) Deep subspace clustering approaches aim to transform the samples into a latent space so that they lie in a union of linear subspaces.}
\label{fig:image_01}
\end{figure}

Recently, deep neural networks have exhibited exceptional ability in capturing complex underlying structures of data and learning discriminative features for clustering \cite{dilokthanakul2016deep,ghasedi2017deep,kazemi2018unsupervised,tian2014learning,xie2016unsupervised}. Inspired by that, a new line of research has been established to bridge deep learning and subspace clustering for developing deep subspace clustering approaches \cite{abavisani2018deep,ji2017deep,peng2016deep,yang2019deep,zhou2018deep}. Variational Autoencoders (VAE) \cite{kingma2013auto,masci2011stacked} and Generative Adversarial Network (GAN) \cite{goodfellow2014generative} are among the most popular deep architectures adopted by these methods to produce feature representations suitable for subspace clustering \cite{masci2011stacked}. Compared to the conventional approaches, deep subspace clustering methods can better exploit the non-linear relationships between the sample points and consequently they achieve superior performance, especially in complex applications in which the samples do not necessarily satisfy the self-expressiveness property \cite{ji2017deep}.

In this paper, we propose a novel spectral clustering-based approach which utilizes stacked convolutional autoencoders to tackle the problem of subspace clustering. Inspired by the idea of residual networks, our first contribution is to add multiple fully-connected linear layers between the corresponding layers of the encoder and decoder to infer multi-level representations from the output of every encoder layer. These connection layers enable to produce representations which are enforced to satisfy self-expressiveness property and hence well-suited to subspace clustering. We model each connection layer as a self-expression matrix created from the summation of a coefficient matrix shared between all layers and a layer-specific matrix that captures the unique knowledge of each individual layer. Moreover, we introduce a novel loss function that utilizes an initial clustering of the samples and efficiently aggregates the information at different levels to infer the coefficient matrix and the layer-specific matrices more accurately. This loss function is further minimized in an iterative scheme which alternatively updates the network parameters for learning better subspace clustering representations and produces a new clustering of the samples. We perform extensive experiments on four benchmark datasets for subspace clustering, including two face image and two object image datasets, to evaluate the efficacy of the proposed method. The experiments demonstrate that our approach can efficiently handle clustering the data from non-linear subspaces and it performs better than the state-of-the-art methods on most of the subspace clustering problems.

\section{Related Works}
Conventional subspace clustering approaches aim to learn a weighted graph whose edge weights represent the relationships between the samples of input data. Then, spectral clustering \cite{ng2002spectral} (or its variants \cite{shi2000normalized}) can be employed to partition the graph into a set of disjoint sub-graphs corresponding to different clusters \cite{chen2009spectral,dyer2013greedy,elhamifar2013sparse,favaro2011closed,heckel2015robust,ji2015shape,liu2012robust,purkait2016clustering,you2016oracle}. A commonly-used formulation to obtain such a weighted graph is written as
\begin{subequations} \label{eq:SSC}
\begin{align}
&\underset{\begin{subarray}{l}
\Cbf\in\Rbb^{n\times n}\end{subarray}}
{\text{minimize}}
&&\frac{1}{2}\lVert\Xbf-\Xbf\Cbf{\rVert}_{\Frm}^2+\lambda\, g(\Cbf) \label{eq:SSC_obj}\\
&\textup{subject to} && \mathrm{diag}(\Cbf)=\mathbf{0},\label{eq:SSC_Con_01}
\end{align}
\end{subequations}
where $\lVert.{\rVert}_{\Frm}$ indicates Frobenius norm, $\Xbf\in\Rbb^{d\times n}$ is a data matrix with its columns representing the samples $\{\xbf_i\in\Rbb^d\}_{i=1}^n$, $\Cbf$ is a self-expression matrix with its $(i,j)^{th}$ element denoting the contribution of sample $\xbf_j$ in reconstructing $\xbf_i$, $g:\Rbb^{n\times n}\to \Rbb$ is a certain regularization function, and $\lambda>0$ is a hyperparameter to balance the importance of the terms. Equality constraint \eqref{eq:SSC_Con_01} is imposed to eliminate the trivial solution $\Cbf=\Ibf_n$ that represents a point as a linear combination of itself. Once the optimal solution $\astaccent{\Cbf}$ of \cref{eq:SSC_obj,eq:SSC_Con_01} is obtained, symmetric matrix $\frac{1}{2}(\lvert\astaccent{\Cbf}\rvert+{\lvert\astaccent{\Cbf}\rvert}^{\!_\top}\!)$ can serve as the affinity matrix of the desired graph where $\lvert.\rvert$ shows the element-wise absolute value operator. Different variants of \cref{eq:SSC_obj,eq:SSC_Con_01} have been well-studied in the literature where they utilize various choices of the regularization function $g(.)$ such as $\lVert\Cbf{\rVert}_{0}$ \cite{yang2016ell,you2016scalable}, $\lVert\Cbf{\rVert}_{1}$ \cite{elhamifar2013sparse}, $\lVert\Cbf{\rVert}_{\ast}$ \cite{luo2018consistent,vidal2014low}, $\lVert\Cbf{\rVert}_{\Frm}$ \cite{peng2013scalable}, etc, to impose desired structures on the graph.

Deep generative architectures, most notably GANs and VAEs, have been widely used in the recent literature to facilitate the clustering task \cite{mukherjee2018clustergan}, especially when the samples come from complex and irregular distributions \cite{masci2011stacked,xie2016unsupervised}. These architectures improve upon the conventional feature extractions by learning more informative and discriminative representations that are highly suitable for clustering \cite{chen2016infogan,peng2018structured,peng2016deep}. To promote inferring clusters with higher quality, some deep approaches propose to jointly learn the representations and perform clustering in a unified framework \cite{mukherjee2018clustergan,peng2017cascade,zhang2019self,zhou2018deep}. One successful deep approach to the subspace clustering problem is presented in \cite{ji2017deep}, known as Deep Subspace Clustering (DSC), which employs a deep convolutional auto-encoder to learn latent representations and uses a novel self-expressive layer to enforce them to lie on a union of linear subspaces. The DSC model is further adopted by Deep Adversarial Subspace Clustering (DASC) method \cite{zhou2018deep} to develop an adversarial architecture, consisting of a generator to produce subspaces and a discriminator to supervise the generator by evaluating the quality of the subspaces. More recently, \cite{zhang2019self} introduced an end-to-end trainable framework, named Self-Supervised Convolutional Subspace Clustering Network (S$^2$ConvSCN), which aims to jointly learn feature representations, self-expression coefficients, and the clustering results to produce more accurate clusters.

Our approach can be seen as a generalization of the DSC algorithm \cite{ji2017deep} to the case that low-level and high-level information of the input data is utilized to produce more informative and discriminative subspace clustering representations. Moreover, we introduce a loss minimization problem that employs an initial clustering of the samples to effectively aggregate the knowledge gained from multi-level representations and to promote learning more accurate subspaces. Notice that although our work is close to DASC \cite{zhou2018deep} and S$^2$ConvSCN \cite{zhang2019self} in the sense that it leverages a clustering of the samples to improve the feature learning procedure, we adopt a completely different strategy to incorporate the pseudo-label information into the problem.

It is noteworthy to emphasize that our approach may seem similar to the multi-view subspace clustering approaches \cite{gao2015multi,luo2018consistent,tang2018learning,zhang2017latent} as it aggregates information obtained from multiple modalities of the data to recover the clusters more precisely. However, it differs from them in the sense that our method leverages some connection layers to simultaneously learn multi-level deep representations and effectively fuse them to boost the clustering performance.

\section{Problem Formulation}
Let $\{\xbf_i\in\Rbb^d\}_{i=1}^n$ be a set of $n$ sample points drawn from a union of $K$ different subspaces in $\Rbb^d$ that are not necessarily linear. An effective approach to cluster the samples is to transform them into a set of new representations that have linear relationships and satisfy the self-expressiveness property. Then, spectral clustering can be applied to recover the underlying clusters. To this end, the DSC algorithm \cite{ji2017deep} introduced a deep architecture consisting of a convolutional autoencoder with $L$ layers to generate latent representations and a fully-connected linear layer inserted between the encoder and decoder to ensure the self-expressiveness property is preserved. Let $\Ecal$ and $\Dcal$, parameterized by $\Theta_{\tav{e}}$ and $\Theta_{\tav{d}}$, denote the encoder and the decoder networks, respectively. Given that, the DSC algorithm proposed to solve the following optimization problem to learn desired representations and infer self-expression matrix $\Cbf$
\begin{subequations} \label{eq:DSC}
\vspace{1mm}
\begin{align}
&\underset{\begin{subarray}{l}
\Theta\end{subarray}}
{\!\text{minimize}}
&&\!\!\!\!\lVert\Xbf\msh-\msh\hat{\Xbf}_{\tav{\Theta}}{\rVert}_{\Frm}^2\msh+\!\lambda\lVert\Zbf_{\tav{\Theta}_{\msh\tavv{e}}}\!\!-\!\Zbf_{\tav{\Theta}_{\msh\tavv{e}}}\!\Cbf{\rVert}_{\Frm}^2\msh+\!\gamma\lVert\Cbf{\rVert}_p \label{eq:DSC_obj}\\
&\textup{\!subject to} &&\!\!\!\! \mathrm{diag}(\Cbf)=\mathbf{0}\label{eq:DSC_Con_01},
\end{align}
\end{subequations}
where $\lambda,\gamma\!>\!0$ are fixed hyperparameters to control the importance of different terms and $\Theta=\{\Theta_{\tav{e}},\Cbf,\Theta_{\tav{d}}\}$ shows the network parameters. Matrix $\Zbf_{\tav{\Theta}_{\tavv{e}}}\in\Rbb^{\bar{d}\times n}$ indicates the latent representations where $\bar{d}$ is the dimension of the representations and $\Zbf_{\tav{\Theta}_{\tavv{e}}}\!\msh=\Ecal(\Xbf;\Theta_{\tav{e}})$, and matrix $\hat{\Xbf}_{\tav{\Theta}}\in\Rbb^{d\times n}$ denotes the reconstructed samples where $\hat{\Xbf}_{\tav{\Theta}}=\Dcal(\Ecal(\Xbf;\Theta_{\tav{e}})\Cbf;\Theta_{\tav{d}})$. The main goal of problem \cref{eq:DSC_obj,eq:DSC_Con_01} is to compute the network parameters such that equality $\Zbf_{\tav{\Theta}_{\msh\tavv{e}}}\!=\Zbf_{\tav{\Theta}_{\msh\tavv{e}}}\!\Cbf$ holds and the reconstructed matrix $\hat{\Xbf}$ can well approximate the input data $\Xbf$. \cite{ji2017deep} used the backpropagation technique followed by the spectral clustering algorithm to find the solution of the minimization problem \cref{eq:DSC_obj,eq:DSC_Con_01} and determine the cluster memberships of the samples. 

In what follows, we propose a new deep architecture that leverages information from different levels of the encoder to learn more informative representations and improve the subspace clustering performance.

\section{Proposed Method}
\begin{figure*}
\centering
\begin{picture}(470,200)
\put(0,0){\includegraphics[width=0.95\textwidth]{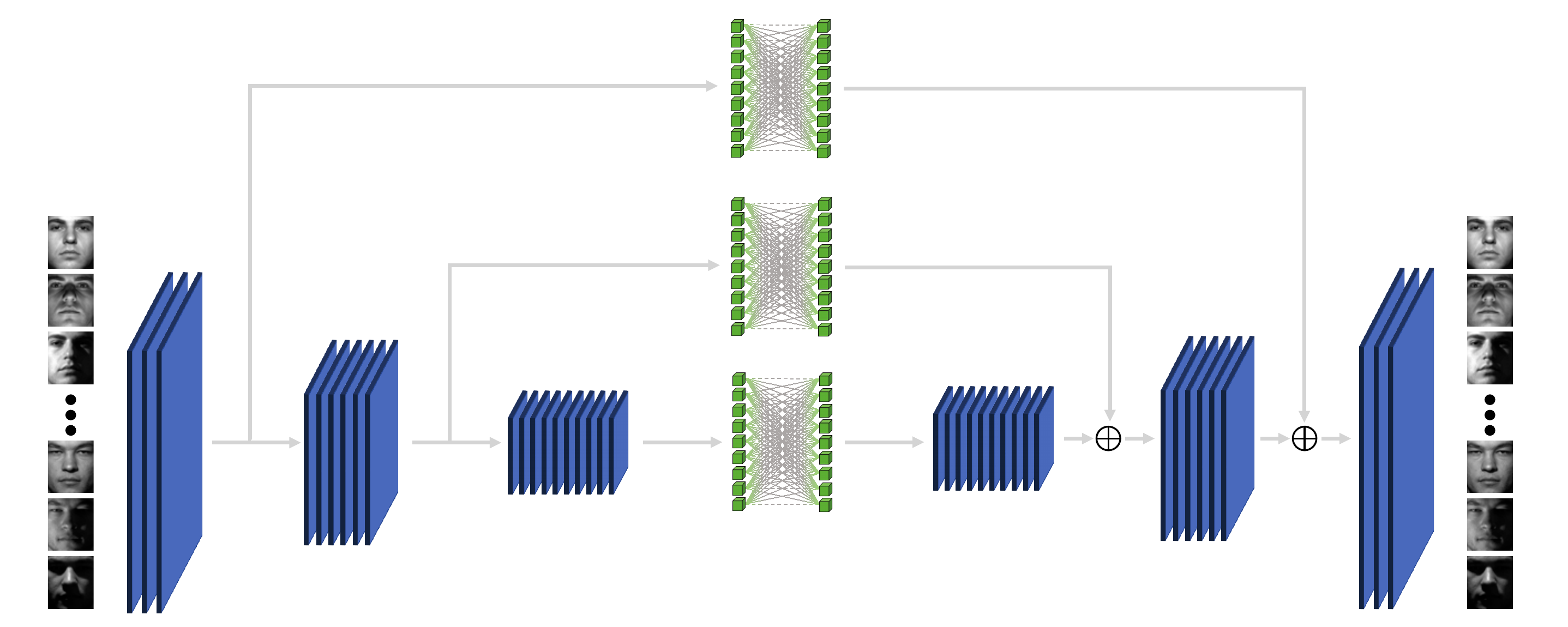}}
\put(2,61){ $\xbf$}
\put(227,29){$\tavvv{\Cbf\msh\msh+\msh\msh\Dbf^{3}}$}
\put(210,33){$\tavvv{\Zbf^{3}}$}
\put(252,33){$\tavvv{\Zbf^{3}(\Cbf\msh\msh+\msh\msh\Dbf^{3})}$}
\put(227,82){$\tavvv{\Cbf\msh\msh+\msh\msh\Dbf^{2}}$}
\put(210,86){$\tavvv{\Zbf^{2}}$}
\put(252,86){$\tavvv{\Zbf^{2}(\Cbf\msh\msh+\msh\msh\Dbf^{2})}$}
\put(227,136){$\tavvv{\Cbf\msh\msh+\msh\msh\Dbf^{1}}$}
\put(210,140){$\tavvv{\Zbf^{1}}$}
\put(252,140){$\tavvv{\Zbf^{1}(\Cbf\msh\msh+\msh\msh\Dbf^{1})}$}
\put(458,61){ $\hat{\xbf}$}
\end{picture}
\caption{Architecture of the proposed multi-level representation learning model
for $L=3$. Observe that the representations learned at different levels of the encoder are fed into fully-connected linear layers to be used in the reconstruction procedure. Such strategy enables to combine low-level information from the early layers with high-level information from the deeper layers to produce more informative and robust subspace clustering representations. Each fully-connected layer is associated with a self-expression matrix formed from the summation of a coefficient matrix $\Cbf$ shared between all layers and a distinctive matrix $\Dbf^{l}$, $l\in\{1,\dots,L\}$, which captures the unique information of each individual layer.}
\label{fig:model}
\end{figure*}
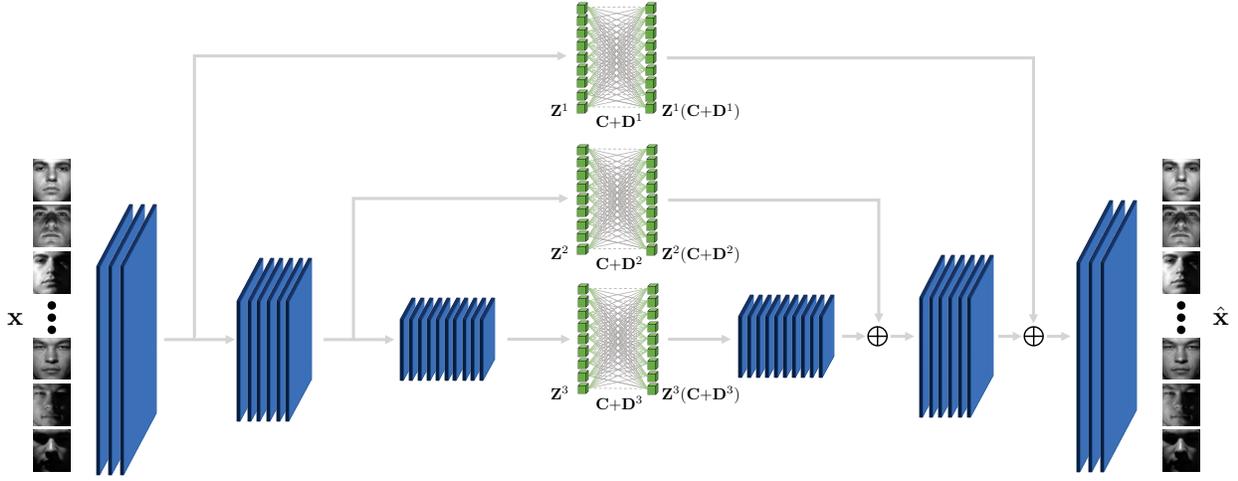

This section presents a detailed explanation of our proposed approach. As it can be seen from the problem \cref{eq:DSC_obj,eq:DSC_Con_01}, the DSC algorithm only relies on the latent variables $\Zbf_{\tav{\Theta}_{\msh\tavv{e}}}$ to perform clustering. Due to the fact that different layers of the encoder provide increasingly complex representations of the input data, it may be quite difficult to learn suitable subspace clustering representations from the output of the encoder. This provides a strong motivation to incorporate information from the lower layers of the encoder to boost the clustering performance. Towards this goal, our approach uses a new architecture which jointly benefits from the low-level and high-level information of the input data to learn more informative subspace clustering representations. The approach adds multiple fully-connected linear layers between the symmetrical layers of the encoder and the decoder to provide multiple paths of information flow through the network. These connection layers can not only enhance the ability of the network in extracting more complex information from the input data but also supervise the output of encoder layers to generate multiple sets of representations that satisfy the self-expressiveness property. \cref{fig:model} illustrates an example architecture of our proposed approach. Observe that the representations learned at different levels of the encoder, denoted as $\{\Zbf_{\tav{\Theta}_{\tavv{e}}}^l\}_{l=1}^L$, are input to the fully-connected linear layers and the outputs of these layers are fed into the decoder layers. This strategy allows the decoder to reuse the low-level information for producing more accurate reconstructions of the input data which in turn can improve the overall clustering performance.

We assume each fully-connected layer is associated with a self-expression matrix in the form of the summation of two matrices, where the first one is shared between the entire layers and the second one is a layer-specific matrix. The encoder, which can be seen as a mapping function from the input space to the representation space, aims to preserve the relations between the data samples at different levels of representations. Moreover, some samples may have stronger (or weaker) relations at different levels of the encoder. Define $\Cbf\in\Rbb^{n\times n}$ as the consistency matrix to capture the relational information shared between the encoder layers and $\{\Dbf^{l}\}_{l=1}^L\in\Rbb^{n\times n}$ as distinctive matrices to produce the unique information of the individual layers. Given that, we incorporate the following loss function to promote learning self-expressive representations
\begin{equation}\label{eq:Loss_Exp}
\begin{aligned}
    \Lcal_{\tav{exp}}=\sum_{l=1}^L\lVert\Zbf_{\tav{\Theta}_{\msh\tavv{e}}}^{l}\!\msh-\!\Zbf_{\tav{\Theta}_{\msh\tavv{e}}}^{l}(\Cbf\msh+\msh\Dbf^{l}){\rVert}_{\Frm}^2.
\end{aligned}
\end{equation}

The above formulation is able to simultaneously model the shared information across different levels while considering the unique knowledge gained from each individual layer. This property allows to effectively leverage the information from the representations learned at multiple levels of the encoder and therefore is also particularly well-suited to the problem of multi-view subspace clustering \cite{luo2018consistent}.

The self-expression loss $\Lcal_{\tav{exp}}$ is employed to promote learning self-expressive feature representations at different levels of the encoder. To better accomplish this purpose, it is beneficial to adopt certain matrix norms for imposing desired structures on the elements of the distinctive matrices $\{\Dbf^{l}\}_{l=1}^L$ and the consistency matrix $\Cbf$. For the distinctive matrices, we use Frobenius norm to ensure the connectivity of the affinity graph associated with each fully-connected layer. For the consistency matrix $\Cbf$, we employ $\ell_1$-norm to generate sparse representations of the data. Ideally, it is desired to infer the consistency matrix and the distinctive matrices such that sample $\xbf_i$ is only expressed by a linear combination of the samples belonging to the same subspace as $\xbf_i$. To ensure these matrices obey the aforementioned desired structures, we propose to incorporate the following regularization terms
\begin{equation} \label{eq:Loss_Reg}
\begin{aligned}
    \Lcal_{\tav{\Cbf}}=\lVert\Qbf^{\!\top}\msh\lvert\Cbf\rvert{\rVert}_{1}, \qquad\Lcal_{\tav{\Dbf}}=\sum_{l=1}^L\lVert\Dbf^{l}{\rVert}_{\Frm}^2,
\end{aligned}
\end{equation}
where $\lVert.{\rVert}_{1}$ computes the sum of absolute values of its input matrix. Regularization term $\Lcal_{\tav{\Cbf}}$ is used to incorporate the information gained from an initial pseudo-labels of the input data into the model. Let $\Qbf\in\Rbb^{n\times K}$ be a membership matrix with its rows are one-hot vectors denoting the initial pseudo-labels assigned to the samples. The multiplication of $\Qbf^{\!\top}$ and $\lvert\Cbf\rvert$ gives a  matrix whose $(i,j)^{th}$ element shows the contribution of the samples assigned to the $i^{th}$ subspace in reconstructing the $j^{th}$ sample. Unlike the commonly used regularization $\lVert\Cbf{\rVert}_{1}$ which imposes sparsity on the entire elements of the consistency matrix $\Cbf$, $\Lcal_{\tav{\Cbf}}$ promotes sparsity on the cluster memberships of the samples. In other words, it encourages each data to be reconstructed by the samples with the same pseudo-label and hence can smooth the membership predictions of the samples to different subspaces. Moreover, the regularization term $\Lcal_{\tav{\Dbf}}$ promotes the elements of the distinctive matrices to be similar in value, which in turn can enhance the connectivity of the affinity graph associated with each fully-connected layer.

Combining the loss function \eqref{eq:Loss_Exp} and the regularization terms $\Lcal_{\tav{\Cbf}}$ and $\Lcal_{\tav{\Dbf}}$ together with the reconstruction loss $\lVert\Xbf-\hat{\Xbf}{\rVert}_{\Frm}^2$ leads to the following optimization problem that needs to be solved for training our proposed model
\begin{subequations} \label{eq:Our}
\vspace{1mm}
\begin{align}
&\underset{\begin{subarray}{l}
\Theta\cup\{\msh\Dbf^l\msh\}_{l=1}^L\end{subarray}}
{\!\text{minimize}}
&&\!\!\!\!\lVert\Xbf\msh-\msh\hat{\Xbf}_{\tav{\Theta}}{\rVert}_{\Frm}^2\msh+\!\lambda_1\msh\sum_{l=1}^L\lVert\Zbf_{\tav{\Theta}_{\msh\tavv{e}}}^{l}\!\msh-\!\Zbf_{\tav{\Theta}_{\msh\tavv{e}}}^{l}(\Cbf\msh+\msh\Dbf^{l}){\rVert}_{\Frm}^2\nonumber\\
&&&+\lambda_2\lVert\Qbf^{\!\top}\msh\lvert\Cbf\rvert{\rVert}_{1}\msh+\msh\lambda_3\msh\sum_{l=1}^L\lVert\Dbf^{l}{\rVert}_{\Frm}^2\!
\label{eq:Our_obj}\\
&\textup{\!subject to} &&\!\!\!\! \mathrm{diag}(\Cbf\msh+\msh\Dbf^{l})=\mathbf{0},\qquad \sm{l\in\{1,\dots,L\}}{9}\label{eq:Our_Con_01},
\end{align}
\end{subequations}
where $\lambda_1,\lambda_2,\lambda_3>0$ are hyperparameters to balance the contribution of different losses. We adopt standard backpropagation technique to obtain the solution of problem \cref{eq:Our_obj,eq:Our_Con_01}. Once the solution matrices $\astaccent{\Cbf}$ and $\{\astaccent{\Dbf}^{l}\}_{l=1}^L$ are obtained, we can create a symmetric affinity matrix $\Wbf\in\Sbb_{n}$ of the following form 
\begin{equation}
\vspace{1mm}
    \begin{aligned}
    \Wbf=\frac{\big\lvert\astaccent{\Cbf}\msh+\msh\frac{1}{L}\sum_{l=1}^L\astaccent{\Dbf}^{l}\big\rvert}{2}\!+\!\frac{\big\lvert\astaccent{\Cbf}^{\!\top}\msh+\msh\frac{1}{L}\sum_{l=1}^L{\astaccent{\Dbf}^{l}}\strut^{\!\top}\big\rvert}{2},
    \end{aligned}
\end{equation}
which shows the pairwise relations between the samples. Given that, the spectral clustering algorithm can be utilized to recover the underlying subspaces and cluster the samples to their respective subspaces. 

Note that the pseudo-labels generated by spectral clustering can be leveraged to retrain the model and provide a more precise estimation of the subspaces. To this end, we assume the membership matrix $\Qbf$ is a variable and develop an iterative scheme to jointly learn the network parameters and matrix $\Qbf$. The approach starts from an initial $\Qbf$ (or equivalently an initial clustering of the input data) and alternatively runs the model for $T$ epochs to train the network parameters $\Theta\cup\{\Dbf^{l}\}_{l=1}^L$ and then updates the membership matrix. This training procedure is then repeated until the number of epochs reaches $\mathrm{maxIter}$. Different steps of our proposed scheme are delineated in detail in \cref{al:Sequential_alg}.  

\begin{algorithm}
\caption{Proposed Subspace Clustering Approach}
\label{al:Sequential_alg}
\begin{algorithmic}[1]
\Require {$\Xbf$, $\Qbf$, $T$, $k=1$}
	\Repeat
		\State \text{Update network parameters $\Theta\cup\{\Dbf^{l}\}_{l=1}^L$ by} \phantom{\;\;\;\;} \text{\phantom{\;\;\;\;\;} solving \cref{eq:Our_obj,eq:Our_Con_01} for one epoch}
		\If {$k$ \text{mod} $T=0$}
		\State \text{Form affinity matrix $\Wbf$}
		\State \text{Apply spectral clustering to update $\Qbf$}
		\EndIf
	    \State $k\leftarrow k+1$
	\Until {$k\leq \mathrm{maxIter}$}
\Ensure {$\Qbf$}
\end{algorithmic}
\end{algorithm}

Observe that \cref{al:Sequential_alg} can train the network parameters $\Theta\cup\{\Dbf^{l}\}_{l=1}^L$ from scratch given the input matrices $\Xbf$, $\Qbf$, and scalar $T$. However, several aspects of the algorithm such as convergence behavior and accuracy can be considerably improved by employing pre-trained models and using fine-tuning techniques to obtain initial values for the encoder and the decoder networks \cite{ji2017deep}. 

In the next section, we perform extensive experiments to corroborate the effectiveness of the proposed model. Also, we present a detailed explanation about the parameter settings, the pre-trained models, and the fine-tuning procedures used in our experiments.

\section{Experiments}
\begin{figure*}
\centering
\scalebox{0.99}{
	\begin{subfigure}[normal]{0.33\linewidth}
		\includegraphics[width=\linewidth]{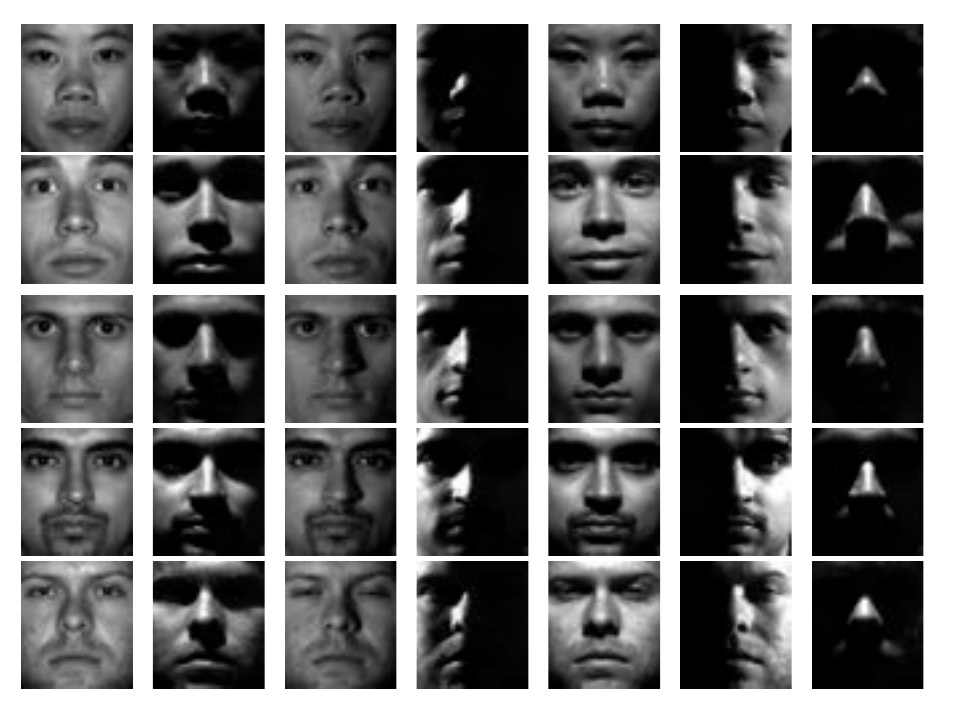}
		\caption{Extended Yale B}
	\end{subfigure}	
	\begin{subfigure}[normal]{0.33\linewidth}
		\includegraphics[width=\linewidth]{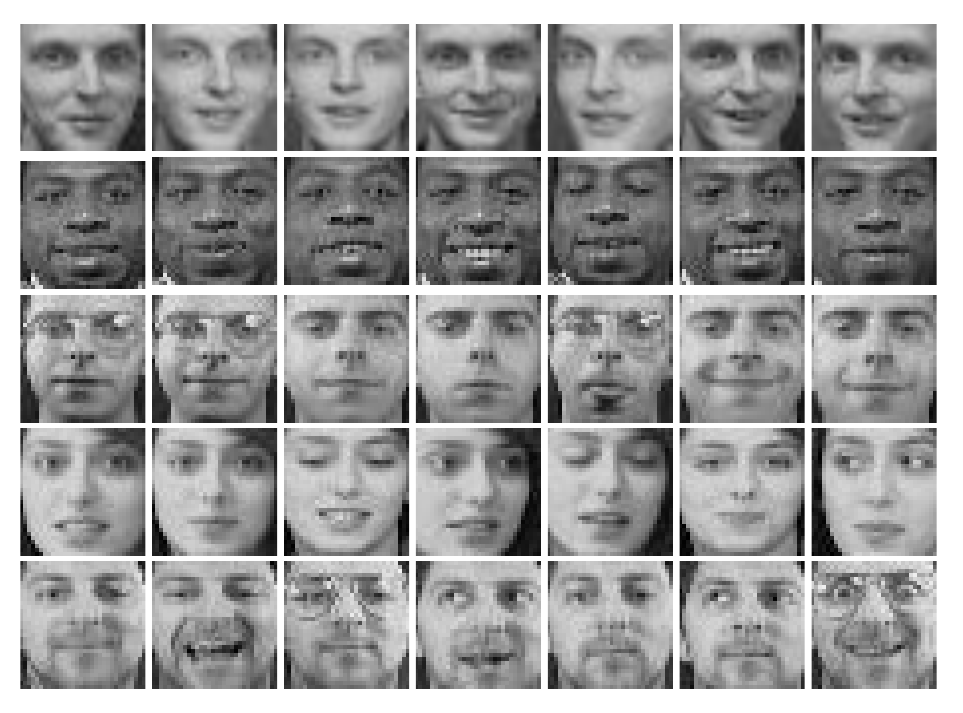}
		\caption{ORL}
	\end{subfigure}
	\begin{subfigure}[normal]{0.33\linewidth}
		\includegraphics[width=\linewidth]{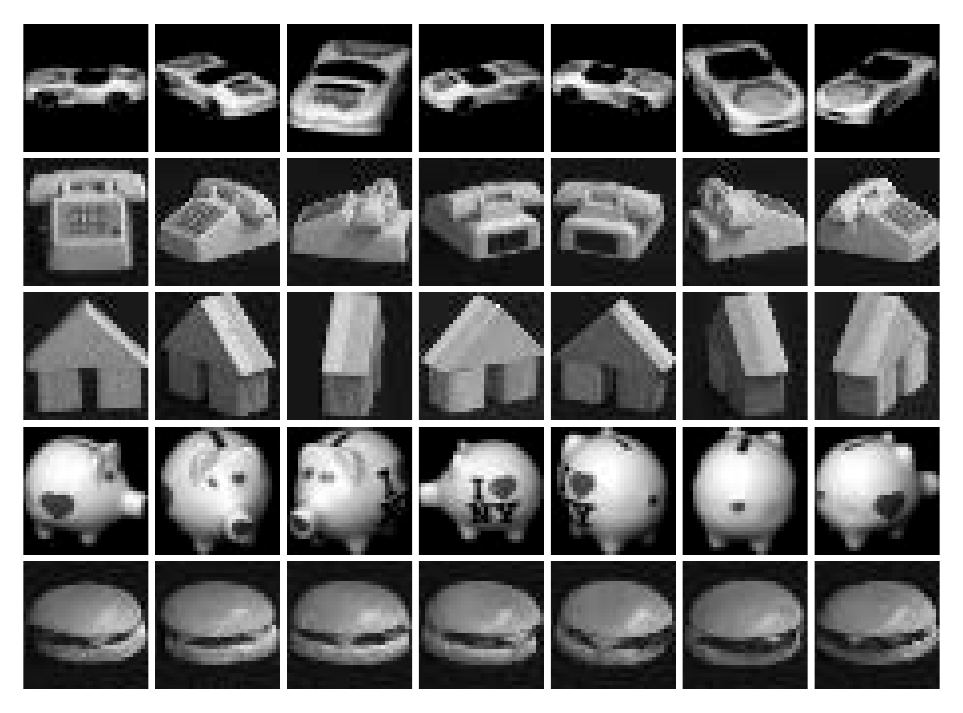}
		\caption{COIL20 and COIL100}
	\end{subfigure}
}
\caption{Example images of Extended Yale B, ORL, COIL20, and COIL100 datasets. The main challenges in the face image datasets, Extended Yale B and ORL, are illumination changes, pose variations and facial expression variations. The main challenges in the object image datasets, COIL20 and COIL100, are the variations in the view-point and scale.}
\label{fig:finalResults1}
\end{figure*}

This section evaluates the clustering performance of our proposed method, termed {\PaperName}, on four standard benchmark datasets for subspace clustering including two face image datasets (ORL and Extended Yale B) and two object image datasets (COIL20 and COIL100). Sample images from each of the datasets are illustrated in \cref{fig:finalResults1}. We perform multiple subspace clustering experiments on the datasets and compare the results against some baseline algorithms, including Low Rank Representation (LRR) \cite{liu2012robust}, Low Rank Subspace Clustering (LRSC) \cite{vidal2014low}, Sparse Subspace Clustering (SSC) \cite{elhamifar2013sparse}, SSC with the pre-trained convolutional auto-encoder features (AE+SSC), Kernel Sparse Subspace Clustering (KSSC) \cite{patel2014kernel}, SSC by Orthogonal Matching Pursuit (SSC-OMP) \cite{you2016scalable}, Efficient Dense Subspace Clustering (EDSC) \cite{ji2014efficient}, EDSC with the pre-trained convolutional auto-encoder features (AE+EDSC),  Deep Subspace Clustering (DSC) \cite{ji2017deep}, and  Deep Adversarial Subspace Clustering (DASC) \cite{zhou2018deep}, Self-Supervised Convolutional Subspace Clustering Network (S$^2$ConvSCN) \cite{zhang2019self}. For the competitor methods, we directly collect the scores from the corresponding papers and some existing literature \cite{ji2017deep,zhang2019self}.

Note that the subspace clustering problem is regarded as a specific clustering scenario which seeks to cluster a set of given unlabeled samples into a union of low-dimensional subspaces that best represent the sample data. In this sense, the subspace clustering approaches are basically different from the standard clustering methods that aim to group the samples around some cluster centers. Most of the subspace clustering literature revolves around using the spectral clustering technique to recover underlying subspaces from an affinity matrix, created over the entire samples. This can considerably increase the computational cost of these methods in comparison to the standard clustering approaches. As a consequence of this limitation, the benchmark datasets used for subspace clustering are generally smaller than that for the clustering task. In this work, we perform experiments on the aforementioned four datasets which are frequently used in the recent literature \cite{elhamifar2013sparse,ji2017deep,zhang2019self,zhou2018deep} to evaluate the performance of the subspace clustering approaches.

In what follows, we first describe the training procedure used in our experiments. Then, we provide more details for each dataset separately and report the clustering performance of state-of-the-art methods.

\subsection{Training Procedure}
Following the literature \cite{ji2017deep,zhou2018deep}, for the convolutional layers, we use kernel filters with stride 2 in both dimensions and adopt rectified linear unit (ReLU) activation function. For the fully-connected layers, we use linear weights without considering bias or non-linear activation function. In order to train the model and obtain the affinity matrix, we follow the literature \cite{ji2017deep,zhang2019self,zhou2018deep} and pass the entire samples into the model as a single batch. The Adam optimizer \cite{kingma2014adam} with $\beta_{1}= 0.9$, $\beta_{2}= 0.999$, and learning rate $0.001$ is used to train the network parameters. All experiments are implemented in PyTorch and the source code will be publicly available on the author's webpage.

As it is mentioned in \cite{ji2017deep}, training the model from scratch is computationally expensive mainly because the samples are passed through the network as a single batch. To address this issue and following \cite{ji2017deep}, we produce a pre-trained model by shortcutting all connection layers (i.e. $\Cbf+\Dbf^{l}=\Ibf$ for $l\in\{1,\dots,L\}$) and ignoring the self-expression loss term $\Lcal_{\tav{exp}}$. The resulting model is trained on the entire sample points and it can be utilized to initialize the encoder and the decoder parameters of our proposed architecture. We initialize the membership matrix $\Qbf$ to a zero matrix in all experiments ($\Lcal_{\tav{\Cbf}}=0$ for the first $T$ epochs), although existing methods can be utilized to obtain better initialization. Moreover, we set each of the individual matrices $\Cbf$ and $\{\Dbf^{l}\}_{l=1}^L$ to a matrix with all elements equal to 0.0001. Notice that \cref{al:Sequential_alg} may fail to generate a convergent sequence of $\Qbf$ as it is terminated after $\mathrm{maxIter}$ epochs. One practical solution to handle this issue is to continue the training procedure until $\Qbf$ converges to a stable matrix \cite{li2015structured}.

\begin{table*}[!t]
\centering
\footnotesize
\caption{Clustering error (\%) of different methods on Extended Yale B dataset. The best results are in bold.}
\scalebox{0.99}{\begin{tabular}{|l||ccccccccccc|}
\hline
Measure      & \text{\scriptsize LRR}   & \text{\scriptsize LRSC}  &  \text{\scriptsize SSC} & \text{\scriptsize AE+SSC}  & \text{\scriptsize KSSC}   & \text{\scriptsize SSC-OMP}& \text{\scriptsize EDSC} & \text{\scriptsize AE+EDSC} & \text{\scriptsize DSC} & \text{\scriptsize S$^2$ConvSCN} & \text{\scriptsize \PaperName}  \\
  \hline
\multicolumn{12}{l}{\vphantom{$1^{1^{.}}_{1}$}\textbf{$\mathbf{10}$ subjects}} 
\\
  \hline
Mean        & $22.22$ & $30.95$ & $10.22$ &  $17.06$ & $14.49$   & $12.08$ & $5.64$ & $5.46$ & $1.59$  & $1.18$ & $\mathbf{1.10}$  \\
Median      & $23.49$ & $29.38$ & $11.09$ &  $17.75$ & $15.78$   & $8.28$  & $5.47$ & $6.09$ & $1.25$  & $1.09$ & $\mathbf{0.94}$  \\ 
  \hline
  \multicolumn{12}{l}{\vphantom{$1^{1^{.}}_{1}$}\textbf{$\mathbf{15}$ subjects}} 
  \\
  \hline
Mean        & $23.22$ & $31.47$ & $13.13$ &  $18.65$ & $16.22$  & $14.05$ & $7.63$ & $6.70$ & $1.69$  & $1.12$ & $\mathbf{0.91}$   \\
Median      & $23.49$ & $31.64$ & $13.40$ &  $17.76$ & $17.34$  & $14.69$ & $6.41$ & $5.52$ & $1.72$  & $1.14$ & $\mathbf{0.99}$   \\
  \hline
  \multicolumn{12}{l}{\vphantom{$1^{1^{.}}_{1}$}\textbf{$\mathbf{20}$ subjects}} 
  \\
  \hline
Mean        & $30.23$ & $28.76$ & $19.75$ &  $18.23$ & $16.55$   & $15.16$ & $9.30$  & $7.67$ & $1.73$ & $1.30$ & $\mathbf{0.99}$   \\
Median      & $29.30$ & $28.91$ & $21.17$ &  $16.80$ & $17.34$   & $15.23$ & $10.31$ & $6.56$ & $1.80$ & $1.25$ & $\mathbf{1.02}$   \\
  \hline
  \multicolumn{12}{l}{\vphantom{$1^{1^{.}}_{1}$}\textbf{$\mathbf{25}$ subjects}} 
  \\
  \hline
Mean        & $27.92$ & $27.81$ & $26.22$ &  $18.72$ & $18.56$   & $18.89$ & $10.67$ & $10.27$  & $1.75$ & $1.29$ & $\mathbf{1.13}$ \\
Median      & $28.13$ & $26.81$ & $26.66$ &  $17.88$ & $18.03$   & $18.53$ & $10.84$ & $10.22$  & $1.81$ & $1.28$ & $\mathbf{1.12}$ \\
  \hline
  \multicolumn{12}{l}{\vphantom{$1^{1^{.}}_{1}$}\textbf{$\mathbf{30}$ subjects}} 
  \\
  \hline
Mean        & $37.98$ & $30.64$ & $28.76$ &  $19.99$ & $20.49$   & $20.75$ & $11.24$ & $11.56$  & $2.07$ & $\mathbf{1.67}$ & $1.78$ \\
Median      & $36.82$ & $30.31$ & $28.59$ &  $20.00$ & $20.94$   & $20.52$ & $11.09$ & $10.36$  & $2.19$ & $1.72$ & $\mathbf{1.41}$ \\
  \hline
  \multicolumn{12}{l}{\vphantom{$1^{1^{.}}_{1}$}\textbf{$\mathbf{35}$ subjects}} 
  \\
  \hline
Mean        & $41.85$ & $31.35$ & $28.55$ &  $22.13$ & $26.07$   & $20.29$ & $13.10$ & $13.28$  & $2.65$ & $1.62$ & $\mathbf{1.44}$ \\
Median      & $41.81$ & $31.74$ & $29.04$ &  $21.74$ & $25.92$   & $20.18$ & $13.10$ & $13.21$  & $2.64$ & $1.60$ & $\mathbf{1.47}$ \\
  \hline
  \multicolumn{12}{l}{\vphantom{$1^{1^{.}}_{1}$}\textbf{$\mathbf{38}$ subjects}} 
  \\
  \hline
Mean        & $34.87$ & $29.89$ & $27.51$ &  $25.33$ & $27.75$   & $24.71$ & $11.64$ & $12.66$   & $2.67$ & $1.52$ & $\mathbf{1.36}$ \\
Median      & $34.87$ & $29.89$ & $27.51$ &  $25.33$ & $27.75$   & $24.71$ & $11.64$ & $12.66$   & $2.67$ & $1.52$ & $\mathbf{1.36}$ \\
 \hline
\end{tabular}}
\label{tab:yaleB}
\end{table*}

\subsection{Results}
The results of all experiments are reported based on the clustering error which is defined to be the percentage of the misclustered samples to the entire sample points.

\vspace{1mm}
\textbf{Extended Yale B:} This dataset is used as a popular benchmark for the subspace clustering problem. It consists of $2432$ frontal face images of size $192\times168$ captured from $38$ different human subjects. Each subject has $64$ images taken under different illumination conditions and poses. For computational purposes and following the literature \cite{elhamifar2013sparse,ji2017deep,zhang2019self}, we downsample the entire images from their original size to $48\times42$. 

We perform multiple experiments for a different number of human subjects $K\in\{10, 15, 20, 25, 30, 35, 38\}$ of the dataset to evaluate the sensitivity of {\PaperName} with respect to increasing the number of clusters. By numbering the subjects from $1$ to $38$, we perform experiments on all possible $K$ consecutive subjects and present the mean and median clustering errors of each $39-K$ trials. Such experiments have been frequently performed in the literature \cite{elhamifar2013sparse,ji2017deep,zhou2018deep,zhang2019self}. Through these experiments, we have employed an autoencoder model consisting of three stacked convolutional encoder layers with $10$, $20$, and $30$ filters of sizes $5\times5$, $3\times3$, and $3\times3$, respectively. The parameters used in the experiments on this dataset are as follows: $\lambda_1=1\times10^{\frac{K}{10}-1}$, $\lambda_2=40$, $\lambda_3=10$, and we update the membership matrix $\Qbf$ in every $T=100$ consecutive epochs. For the entire choices of $K$, we set the maximum number of epochs to $900$. The clustering results on this dataset are reported in \cref{tab:yaleB}. Observe that {\PaperName} achieves smaller errors than the competitor methods in all experiments, except for the mean of clustering error in case $K=30$. 

\vspace{1mm}
\noindent\textbf{ORL:} This dataset consists of $400$ face images of size $112\times 92$ from $40$ different human subjects where each subject has $10$ images taken under diverse variation of poses, lighting conditions, and facial expressions. Following the literature, we downsample the images from their original size to $32\times32$. This dataset is challenging for subspace clustering due to the large variation in the appearance of facial expressions (shown in \cref{fig:finalResults1}) and since the number of images per each subject is quite small.

Through the experiment on ORL, we have adopted a network architecture consisting of three convolutional encoder layers with $3$, $3$, and $5$ filters, all of size $3\times 3$. Moreover, the parameter settings used in the experiment are as follows: $\lambda_1=5$, $\lambda_2=0.5$, $\lambda_3=1$, $T=10$, and the maximum number of epochs is set to $420$. The results of this experiment are presented in \cref{tab:COILORL}. It can be seen that {\PaperName} outperforms all the competitor methods, except S$^2$ConvSCN which attains the smallest clustering error rate on ORL.

\vspace{1mm}
\noindent\textbf{COIL20/COIL100}: These two datasets are widely used for different types of clustering. COIL20 contains $1440$ images captured from $20$ various objects and COIL100 has $7200$ images of $100$ objects. Each object in either of the datasets has $72$ images with black background taken at pose intervals of $5$ degrees. The large viewpoint changes can pose serious challenges for the subspace clustering problem on these two dataset (Shown in in \cref{fig:finalResults1}).
\begin{table*}[!t]
\centering
\footnotesize
\caption{Clustering error (\%) of different methods on ORL, COIL20, and COIL100 datasets. The best results are in bold.}
\scalebox{0.99}{\begin{tabular}{|c||cccccccccccc|}
\hline
   Dataset  & \text{\scriptsize LRR}   & \text{\scriptsize LRSC}  &  \text{\scriptsize SSC} & \text{\scriptsize AE+SSC}  & \text{\scriptsize KSSC}   & \text{\scriptsize SSC-OMP}& \text{\scriptsize EDSC} & \text{\scriptsize AE+EDSC} & \text{\scriptsize DSC} & \text{\scriptsize DASC} & \text{\scriptsize S$^2$ConvSCN} & \text{\scriptsize \PaperName}  \\
  \hline
  \hline
   ORL    & $33.50$  & $32.50$  & $29.50$  &  $26.75$  & $34.25$  & $37.05$  & $27.25$   &  $26.25$  & $14.00$  & $11.75$  & $\mathbf{10.50}$  &  $11.25$ \\
  \hline
  \hline
   COIL20  & $30.21$  & $31.25$  & $14.83$  &  $22.08$   & $24.65$  & $29.86$  & $14.86$  & $14.79$   & $5.42$   & $3.61$   &  $2.14$  &  $\mathbf{2.08}$ \\
  \hline
  \hline
    COIL100  & $53.18$  & $50.67$  & $44.90$  &  $43.93$    & $47.18$  & $67.29$  & $38.13$  & $38.88$    & $30.96$  & $-$    &  $26.67$ &  $\mathbf{23.28}$ \\
 \hline
\end{tabular}}
\label{tab:COILORL}
\end{table*}
\begin{table}[!t]
\centering
\footnotesize
\caption{Ablation study of our method in terms of clustering error (\%) on Extended Yale B. The best results are in bold.}
\scalebox{0.99}{\begin{tabular}{|l||cccc|}
\hline
  Measure      & \text{\scriptsize DSC-L$2$} &  \text{\scriptsize DSC-L$1$} & \text{\scriptsize MLRDSC ($\lVert\Cbf{\rVert}_{1}$)} & \text{\scriptsize MLRDSC}    \\
  \hline
  \multicolumn{4}{l}{\vphantom{$1^{1^{.}}_{1}$}\textbf{$\mathbf{10}$ subjects}} 
  \\
  \hline
  Mean        & $1.59$ & $2.23$ & $\mathbf{1.09}$ &  $1.10$  \\
  Median      & $1.25$ & $2.03$ & $1.08$ &        $\mathbf{0.94}$   \\
  \hline
  \multicolumn{4}{l}{\vphantom{$1^{1^{.}}_{1}$}\textbf{$\mathbf{15}$ subjects}} 
  \\
  \hline
  Mean        & $1.69$ & $2.17$ & $0.98$          &  $\mathbf{0.91}$   \\
  Median      & $1.72$ & $2.03$ & $\mathbf{0.99}$ &  $\mathbf{0.99}$   \\
  \hline
  \multicolumn{4}{l}{\vphantom{$1^{1^{.}}_{1}$}\textbf{$\mathbf{20}$ subjects}} 
  \\
  \hline
  Mean        & $1.73$ & $2.17$ & $\mathbf{0.94}$ &  $0.99$   \\
  Median      & $1.80$ & $2.11$ & $\mathbf{0.94}$ &  $1.02$   \\
  \hline
  \multicolumn{4}{l}{\vphantom{$1^{1^{.}}_{1}$}\textbf{$\mathbf{25}$ subjects}} 
  \\
  \hline
  Mean        & $1.75$ & $2.53$ & $\mathbf{1.13}$ &  $\mathbf{1.13}$  \\
  Median      & $1.81$ & $2.19$ & $\mathbf{1.12}$ &  $\mathbf{1.12}$  \\
  \hline
  \multicolumn{4}{l}{\vphantom{$1^{1^{.}}_{1}$}\textbf{$\mathbf{30}$ subjects}} 
  \\
  \hline
  Mean        & $2.07$ & $2.63$ & $1.84$          & $\mathbf{1.78}$  \\
  Median      & $2.19$ & $2.81$ & $\mathbf{1.35}$ & $1.41$  \\
  \hline
  \multicolumn{4}{l}{\vphantom{$1^{1^{.}}_{1}$}\textbf{$\mathbf{35}$ subjects}} 
  \\
  \hline
  Mean        & $2.65$ & $3.09$ & $1.49$ &  $\mathbf{1.44}$  \\
  Median      & $2.64$ & $3.10$ & $1.49$ &  $\mathbf{1.47}$  \\
  \hline
  \multicolumn{4}{l}{\vphantom{$1^{1^{.}}_{1}$}\textbf{$\mathbf{38}$ subjects}} 
  \\
  \hline
  Mean        & $2.67$ & $3.33$ & $1.40$ &  $\mathbf{1.36}$  \\
  Median      & $2.67$ & $3.33$ & $1.40$ &  $\mathbf{1.36}$  \\
 \hline
\end{tabular}}
\label{tab:AblationStudy}
\end{table}

For COIL20 and COIL100 datasets, the literature methods \cite{ji2017deep,zhang2019self,zhou2018deep} mostly adopt one layer convolutional autoencoders to learn feature representations. This setting admits no connection layer and hence is not well-suited to our approach. To better demonstrate the advantages of {\PaperName}, we use a two layers convolutional autoencoder model with $5$ and $10$ filters for performing experiment on COIL20 and adopt the same architecture with $20$ and $30$ filters for COIL100. The entire filters used in both experiments are of size $3\times 3$. Moreover, the parameter settings for the datasets are as follows: 1)\,COIL20: $\lambda_1=20$, $\lambda_2=20$, $\lambda_3=5$, $T=5$, and the maximum number of epochs is set to $50$; 2)\,COIL100: $\lambda_1=20$, $\lambda_2=40$, $\lambda_3=10$, $T=50$, and the maximum number of epochs is set to $350$. The results on COIL20 and COIL100 datasets are shown in \cref{tab:COILORL}. Observe that our approach achieves better subspace clustering results on both datasets compared to the existing state-of-the-art methods.

\begin{table}[!t]
\centering
\footnotesize
\caption{Sensitivity analysis of our method in terms of clustering error (\%) on Extended Yale B. Triplet ($\bar{\lambda}_1$, $\bar{\lambda}_2$, $\bar{\lambda}_3$) corresponds to the parameter setting used to produce the results of \cref{tab:yaleB}.}
\scalebox{1.05}{
\begin{tabular}{|l||@{\;\;}c@{\;\;\;}c@{\;\;\;}c@{\;\;\;}c@{\;\;\;}c@{\;\;\;}c@{\;\;\;}c@{\;\;}|}
\hline
  \multicolumn{1}{|c||}{$\lambda_{1}^{\phantom{1^2}}$} & $\bar{\lambda}_1$ & $\bar{\lambda}_1$ & $\bar{\lambda}_1$ & $\bar{\lambda}_1$ & $\bar{\lambda}_1$ & $0.1 \bar{\lambda}_1$ & $10\bar{\lambda}_1$ \\
  \multicolumn{1}{|c||}{$\lambda_{2}^{\phantom{1^2}}$} & $\bar{\lambda}_2$ & $0.1\bar{\lambda}_2$ & $100\bar{\lambda}_2$ & $\bar{\lambda}_2$ & $\bar{\lambda}_2$ & $\bar{\lambda}_2$ & $\bar{\lambda}_2$ \\
  \multicolumn{1}{|c||}{$\lambda_{3}^{\phantom{1^2}}$} & $\bar{\lambda}_3$ & $\bar{\lambda}_3$ & $\bar{\lambda}_3$ & $0.1\bar{\lambda}_3$ & $100\bar{\lambda}_3$ & $\bar{\lambda}_3$ & $\bar{\lambda}_3$ \\
  \hline
  \multicolumn{4}{l}{\vphantom{$1^{1^{.}}_{1}$}\textbf{$\mathbf{10}$ subjects}} 
  \\
  \hline
  Mean & 1.10 & 1.11 & 1.15 &  1.09 & 1.06 & 2.52 & 1.08 \\
  Median & 0.94 & 0.94 & 0.94 & 0.94 & 0.94 & 2.03 & 1.09 \\
  \hline
  \multicolumn{4}{l}{\vphantom{$1^{1^{.}}_{1}$}\textbf{$\mathbf{15}$ subjects}} 
  \\
  \hline
  Mean & 0.91 & 0.92 & 0.99 &  0.92 & 0.94 & 1.28 & 0.97 \\
  Median & 0.99 & 0.99 & 1.04 & 0.99 & 0.99 & 1.25 & 0.94 \\
  \hline
  \multicolumn{4}{l}{\vphantom{$1^{1^{.}}_{1}$}\textbf{$\mathbf{20}$ subjects}} 
  \\
  \hline
  Mean & 0.99 & 0.99 & 1.00 & 0.99 & 1.00 & 0.83 & 0.94 \\
  Median & 1.02 & 1.02 & 1.02 & 1.02 & 1.02 & 0.86 & 1.02 \\
  \hline
  \multicolumn{4}{l}{\vphantom{$1^{1^{.}}_{1}$}\textbf{$\mathbf{25}$ subjects}} 
  \\
  \hline
  Mean & 1.13 & 1.13 & 1.14 & 1.13 & 1.14 & 1.51 & 1.13 \\
  Median & 1.12 & 1.09 & 1.13 & 1.12 & 1.09 & 1.06 & 1.13 \\
  \hline
  \multicolumn{4}{l}{\vphantom{$1^{1^{.}}_{1}$}\textbf{$\mathbf{30}$ subjects}} 
  \\
  \hline
  Mean & 1.78 & 2.26 & 2.42 & 1.86 & 1.97 & 2.43 & 1.70 \\
  Median & 1.41 & 1.35 & 1.41 & 1.41 & 1.41 & 1.46 & 1.35 \\
  \hline
  \multicolumn{4}{l}{\vphantom{$1^{1^{.}}_{1}$}\textbf{$\mathbf{35}$ subjects}} 
  \\
  \hline
  Mean & 1.44 & 1.45 & 1.45 & 1.45 & 1.45 & 1.45 & 1.50 \\
  Median & 1.47 & 1.47 & 1.47 & 1.47 & 1.47 & 1.47 & 1.50 \\
  \hline
  \multicolumn{4}{l}{\vphantom{$1^{1^{.}}_{1}$}\textbf{$\mathbf{38}$ subjects}} 
  \\
  \hline
  Mean & 1.36 & 1.36 & 1.32 & 1.36 & 1.36 & 1.32 & 1.40 \\
  Median & 1.36 & 1.36 & 1.32 & 1.36 & 1.36 & 1.32 & 1.40 \\
 \hline
\end{tabular}
}
\label{tab:Sensitivity}
\end{table}

According to the \cref{tab:yaleB,tab:COILORL}, the deep subspace clustering methods, such as DSC, S$^2$ConvSCN, and {\PaperName}, perform considerably well compared to the classical subspace clustering approaches on the benchmark datasets. This success can be attributed to the fact that deep models are able to efficiently capture the non-linear relationships between the samples and recover the underlying subspaces. Moreover, the results indicate that {\PaperName} outperforms the DSC algorithm by a notable margin. This improvement can be resulted from the incorporation of a modified regularization term and the insertion of connection layers between the corresponding layers of the encoder and decoder. These layers enable the model to combine the information of different levels of the encoder to learn more favorable subspace clustering representations. It is noteworthy to mention that although our approach achieves better clustering results than the DSC method, it has more parameters to train, which in turn increases the computational burden of the model.

\vspace{1mm}
\noindent\textbf{Ablation Study:}
To highlight the benefits brought by different components of our proposed model, we carry out an ablation study by evaluating a variant of our approach, named {\PaperName} ($\lVert\Cbf{\rVert}_1$), which replaces $\Lcal_{\tav{\Cbf}}$ with $\lVert\Cbf{\rVert}_1$. In this sense, {\PaperName} ($\lVert\Cbf{\rVert}_1$) can be seen as a generalization of DSC-L$1$ (a variant of the DSC algorithm that utilizes regularization term $\lVert\Cbf{\rVert}_1$ \cite{ji2017deep}) to a case that leverages multiple connection layers to learn multi-level subspace clustering representations. We perform experiments for different number of subjects $K$ on Extended Yale B dataset and present the clustering results in \cref{tab:AblationStudy}. As the table indicates, inserting the connection layers between the symmetrical layers of the encoder and decoder can considerably improve the clustering performance of DSC-L$1$ algorithm. Moreover, comparing the results of {\PaperName} and {\PaperName}($\lVert\Cbf{\rVert}_1$) confirms the positive effect of incorporating the regularization term $\Lcal_{\tav{\Cbf}}$.

\vspace{1mm}
\noindent\textbf{Sensitivity Analysis:} we perform multiple experiments on the Extended Yale B dataset with various choices of hyperparameters ($\lambda_1$, $\lambda_2$, $\lambda_3$) to evaluate the sensitivity of the proposed approach to the choice of these parameters. The results of these experiments are reported in \cref{tab:Sensitivity}. Observe that the proposed approach exhibits a satisfactory performance for a wide range of these hyperparameters which demonstrates its generalization power.

\section{Conclusions}
This paper presented a novel spectral clustering-based approach which uses a deep neural network architecture to address subspace clustering problem. The proposed method improves upon the existing deep approaches by leveraging information exploited from different levels of the networks to transform input samples into multi-level representations lying on a union of linear subspace. Moreover, it is able to use pseudo-labels generated by spectral clustering technique to effectively supervise the representation learning procedure and boost the final clustering performance. Experiments on benchmark datasets demonstrate that the proposed approach is able to efficiently handle clustering from the non-linear subspaces and it achieves better results compared to the state-of-the-art methods.


{\small
\bibliographystyle{ieee}
\bibliography{egbib}
}

\end{document}